\documentclass[letterpaper]{article} % DO NOT CHANGE THIS
\usepackage[]{aaai23}  % DO NOT CHANGE THIS
\usepackage{times}  % DO NOT CHANGE THIS
\usepackage{helvet}  % DO NOT CHANGE THIS
\usepackage{courier}  % DO NOT CHANGE THIS
\usepackage{amssymb}
\usepackage[hyphens]{url}  % DO NOT CHANGE THIS
\usepackage{graphicx} % DO NOT CHANGE THIS
\usepackage{natbib}  % DO NOT CHANGE THIS AND DO NOT ADD ANY OPTIONS TO IT
\usepackage{caption} % DO NOT CHANGE THIS AND DO NOT ADD ANY OPTIONS TO IT
\usepackage{algorithm}
\usepackage{algorithmic}
\usepackage{xspace}
\usepackage{amsfonts} 
\newcommand{\down}{}
\usepackage{amsmath}
\usepackage{multirow}
\usepackage{rotating,tabularx}
\usepackage{booktabs}
\usepackage{listings}
\usepackage[np, autolanguage]{numprint}
\usepackage{paralist}
\usepackage{enumitem}
\usepackage{colortbl}
\usepackage{cancel}
\usepackage{soul}
\usepackage{epstopdf}
\usepackage{pifont}
\usepackage{xcolor}
\usepackage{mathrsfs}
\usepackage{todonotes}
\usepackage{subfig}
\newcommand{\ie}{\textit{i.e.,}\xspace}

\definecolor{bleudefrance}{rgb}{0.19, 0.55, 0.91}
\definecolor{yes}{RGB}{239,211,69}
\definecolor{carminered}{rgb}{1.0, 0.0, 0.22}
\definecolor{crimsonglory}{rgb}{0.75, 0.0, 0.2}

\usepackage{newfloat}
\usepackage{listings}
\DeclareCaptionStyle{ruled}{labelfont=normalfont,labelsep=colon,strut=off} % DO NOT CHANGE THIS
\floatstyle{ruled}
\newfloat{listing}{tb}{lst}{}
\floatname{listing}{Listing}
\title{Learning towards Selective Data Augmentation for Dialogue Generation}
\author{
    Xiuying Chen\textsuperscript{\rm 1}\equalcontrib,
    Mingzhe Li\textsuperscript{\rm 2}\equalcontrib,
   Jiayi Zhang\textsuperscript{\rm 3},\\
   Xiaoqiang Xia\textsuperscript{\rm 3},
    Chen Wei\textsuperscript{\rm 3},
    Jianwei Cui\textsuperscript{\rm 3},
   Xin Gao\textsuperscript{\rm 1$\dag$}, Xiangliang Zhang\textsuperscript{\rm 4},
    Rui Yan\textsuperscript{\rm 5}\footnote{Corresponding authors.}
}
\affiliations{
    \textsuperscript{\rm 1}Computational Bioscience Research Center, KAUST\\
    \textsuperscript{\rm 2}Ant Group\\ \textsuperscript{\rm 3} Xiaomi AI Lab\\ 
    \textsuperscript{\rm 4} University of Notre Dame \\
    \textsuperscript{\rm 5}Gaoling School of Artificial Intelligence, Renmin University of China
    xiuying.chen@kaust.edu.sa,limingzhe.lmz@antgroup.com
}

\usepackage{bibentry}
\begin{document}

\maketitle

\begin{abstract}
% Current state-of-the-art neural dialog models learned from human conversations are based on the data-driven paradigm.
As it is cumbersome and expensive to acquire a huge amount of data for training neural dialog models, data augmentation is proposed to effectively utilize existing training samples.
However, current data augmentation techniques on the dialog generation task mostly augment all cases in the training dataset without considering the intrinsic attributes between different cases.
We argue that not all cases are beneficial for augmentation task, and the cases suitable for augmentation should obey the following two attributes: 
(1) low-quality (the dialog model cannot generate a high-quality response for the case),
(2) representative (the case should represent the property of the whole dataset).
Herein, we explore this idea by proposing a \textit{Selective Data Augmentation framework} (SDA) for the response generation task.
SDA employs a dual adversarial network to select the lowest quality and most representative data points for augmentation in one stage. 
Extensive experiments conducted on two publicly available datasets, \ie DailyDialog and OpenSubtitles, show that our framework can improve the response generation performance with respect to various metrics.
\end{abstract}

	\section{Introduction}
	\label{intro}

Open-domain dialogue generation is becoming a research hotspot in the community of natural language processing due to its penitential applications \cite{Li2019InsufficientDC,chen2021reasoning}. 
	Generally, in the paradigm of deep neural networks, a large quantity of training data is required for facilitating the convergence of these models.
	As such, a data augmentation framework that can generate reliable training cases is the crux of building a robust dialogue model.
	
\begin{figure}[tb]
		\centering
		\includegraphics[width=1\linewidth]{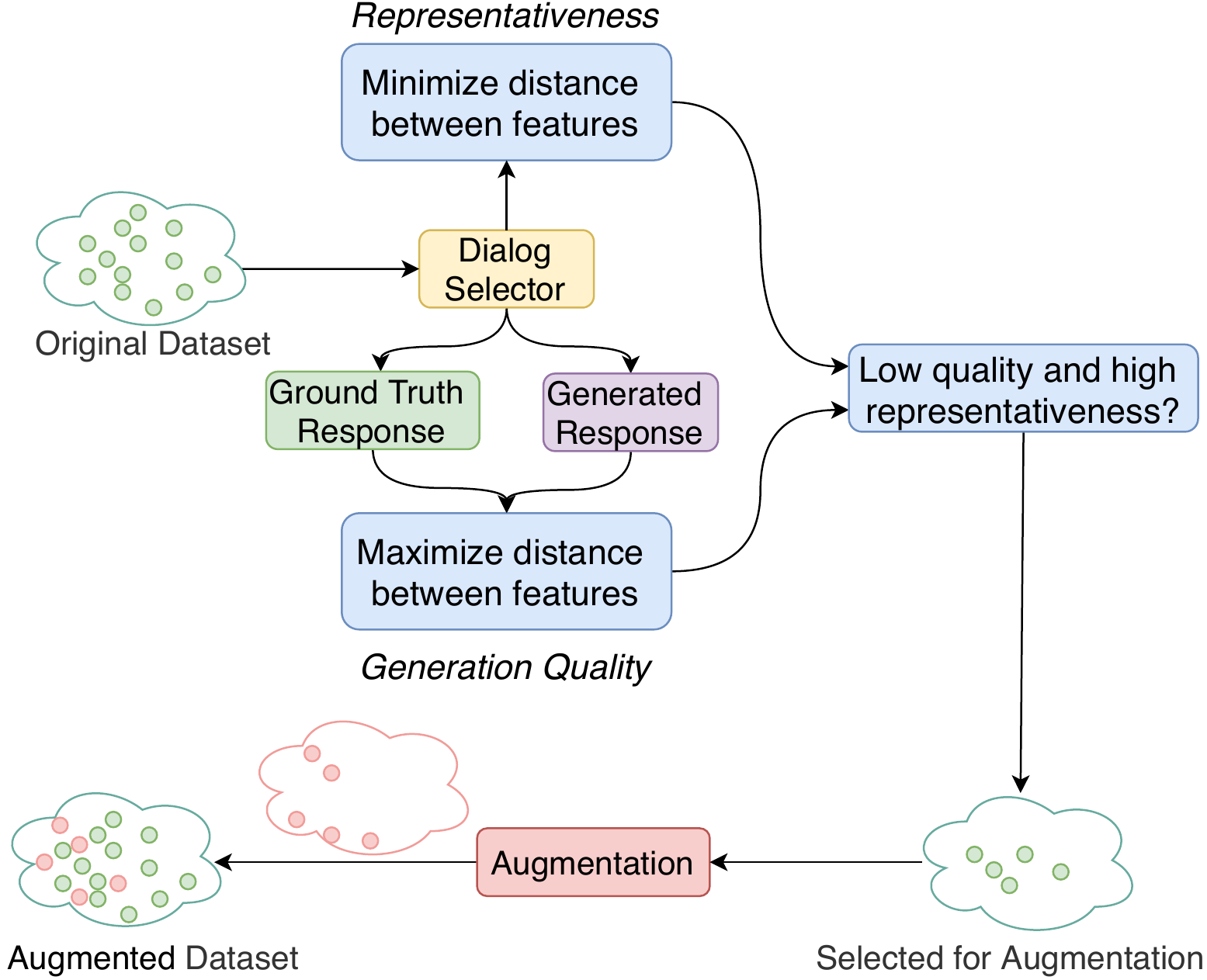}
		\caption{Our goal is to simultaneously select the lowest quality and the most representative cases in the training dataset for augmentation (best viewed in color).}
% 		The dialog selector is required to select the samples maximizing the distance between generated responses and original responses (low-quality), while minimizing the distance between selected samples and original samples (representative).}
		\label{fig:guidance}
	\end{figure}

	As shown in Figure\ref{fig:guidance}(a), existing data augmentation methods for the dialog generation task mainly investigate different ways to augment all data samples without considering their distinct attributes.
	For example,
	\citet{Hou2018SequencetoSequenceDA} augmented each case by leveraging other cases with similar semantic meaning in the training dataset, and ~\citet{Li2019InsufficientDC} generated diversified versions for each query and response in an adversarial style. 
	However, we argue that in practice, the attributes of the training cases vary, thus, not all cases are necessary for augmentation.
    The augmentation of dull responses such as ``I don't know'' and noisy samples with unpaired queries and responses even brings harm to the model.
	Taking one step further, we assume that whether each case is beneficial for augmentation should be examined from two aspects.
	From the generation quality aspect, the generation model may perform relatively well in some cases, for example, the cases with safe answers.
	Correspondingly, it is redundant and sometimes harmful to augment these cases \cite{csaky-etal-2019-improving}.
	Thus, we should only focus on part of the data where the model fails to adapt to (\textit{low-quality}).
	From the dataset attribute side, the quality of user-generated training data varies greatly, and noisy samples frequently appear~\cite{Cai2020DataMT}.
	Hence, we should augment the representative cases that reflect a larger set of their properties (\textit{representative}), instead of some noisy samples that do not represent the general attribute of the whole dataset.
	This is also inspired by a previous study \cite{Schrder2020ASO}, which shows that training on representative cases can increase the quality of the resulting model.

	Based on this assumption, in this work, we propose a novel \textit{Selective Data Augmentation} framework, namely SDA, to accurately select the most informative data points from the training dataset by simultaneously considering the generation quality and representativeness.
	The overview is illustrated in Figure~\ref{fig:guidance}(b). 
	The dialog selector is required to select the samples maximizing the distance between generated responses and original responses (low-quality) while minimizing the distance between selected samples and original samples (representative).

	Concretely, we use a dual generative adversarial (Dual-GAN) framework to assist the dialog selector in the distance measurement between deep feature representations.
	From the generation quality side, a discriminator tries to discriminate between the generated response and the ground-truth response, while the dialog selector aims to trick the discriminator.
	If the generated responses cannot fool the discriminator, then the selected samples have low quality.
	From the representativeness side, we measure the distance by the reconstruction process.
	If the selected samples successfully reconstruct the original data, then the selected cases have high representativeness.
Concretely, the samples selected by the dialog selector are sent to a variational autoencoder (VAE), which embeds the features of selected samples into the same latent space and then reconstructs them. 
	The reconstructed features are fed to the representativeness discriminator, which tries to discriminate between the original samples and the reconstructed samples.
	If the selected samples successfully fool the discriminator, then the selected samples have high representativeness.
	In this way, the dialog selector is encouraged to take both generation quality and representativeness into consideration during data selection.

	Our main contributions can be summarized as follows:
	(1) We propose the selective data augmentation task, which aims to select suitable training cases for augmentation.
	(2) We propose a dual adversarial framework for \textit{Selective Data Augmentation} (SDA), which can simultaneously learn to select the lowest quality and most representative data points for augmentation.
	(3) Extensive experiments conducted on two public dialog datasets show that our approach can improve the dialog generation performance. 
	We also show the universality of our framework for the story generation task.

	\section{Related Work}
% 	In this section, we summarize the related work about dialog generation and data augmentation.
	
	\textbf{Dialog Generation.}
	Existing approaches to improve neural dialogue generation models mainly target building more powerful learning systems, using extra information such as conversation topics \cite{Zou2020TopicOrientedSD}, persona profile \cite{Chan2019ModelingPI}, user emotions \cite{song2019generating}, out-sourcing knowledge \cite{li2021stylized}, or pretrained models \cite{tuan2021local}. 
	Another popular framework for dialogue generation concentrates on using VAE \cite{Zhao2017LearningDD}, in which a latent variable is introduced to benefit the dialogue model with more diverse response generation.
	As the GAN framework facilitates training the generator,
% 	\citet{Larsen2016AutoencodingBP} proposed to integrate VAE and GAN, where VAE is treated as the generator.
	% \citet{Hu2017TowardCG} first combined VAE and GAN for text generation. 
	% In this work, we investigate employing VAE and GAN to learn the attributes of training samples for data augmentation in dialog generation tasks.

	\textbf{Data Augmentation.}
	In the paradigm of deep learning, data augmentation is an effective way to boost the performance of neural models.
	To name a few, \citet{Kurata2016LabeledDG} proposed to generate labeled data with the decoder LSTM based on the perturbated encoded vector for the semantic slot filling task.
	% \citet{Sennrich2016ImprovingNM} boosted neural machine translation models using back-translation.
\citet{Andreas2020GoodEnoughCD} designed a simple data augmentation protocol aimed at providing a compositional inductive bias in conditional and unconditional sequence models.
	\citet{Kobayashi2018ContextualAD} and \citet{Wu2019ConditionalBC} showed that contextual augmentation using label-conditional language models helps text classification tasks.
	In terms of dialog generation task, 	\citet{Li2019InsufficientDC} proposed a generative model to augment existing data, where the CVAE was employed as the generator to output more training data with diversified expressions.
\citet{louvan2020simple} proposed Lightweight augmentation, a set of word-span and sentence-level augmentation methods for low-resource slot filling and intent classification.
	The most recent work was proposed by \citet{Cai2020DataMT}, where they proposed to augment and reweight all learning samples.
	% In our work, we propose to selectively augment training cases instead of augmenting all.
	
	\section{Methodology}
	\label{sec:formulation}

%	To formulate our task, the dialogue generation and data augmentation processes are described with necessary notations, shown as follows.

	Open-domain dialogue generation involves generating a response $R^i = (r^i_1,...,r^i_j,...,r^i_m)$ for a user-issued query $Q^i = (q^i_1,...,q^i_k,...,q^i_{m'})$, where $r^i_j$ refers to the $j$-th word in the response in $i$-th case, and $q^i_k$ denotes the $k$-th word in the query in  $i$-th case.
	$m$ and $m'$ are the word length of a response and a query, respectively. 
	The entire dialogue
	system is trained under $D$, \ie maximizing the $P (R^i | Q^i)$, where $D =\{(Q^i , R^i )\}^N_{i=1}$ is the dataset and $N$ refers to the number of training query-response pairs. 
	For the data augmentation task, the original dataset $D$ is increased to $D' =\{(Q^i , R^i )\}^{N'}_{i=1}$, where $N'$ is the data size after augmentation.
	In our selective data augmentation task, we aim to select suitable cases suitable for augmentation and increase the data size from $N$ to $N'$.
	Correspondingly, the response generation changes from $\text{argmax}P(R|Q,D)$ to $\text{argmax} P(R|Q,D')$.

% 	In this section, we propose our \textit{Selective Data Augmentation} framework (SDA), which is shown in Figure~\ref{fig:model}.
	
\begin{figure*}[tb]
		\centering
		\includegraphics[width=0.75\linewidth]{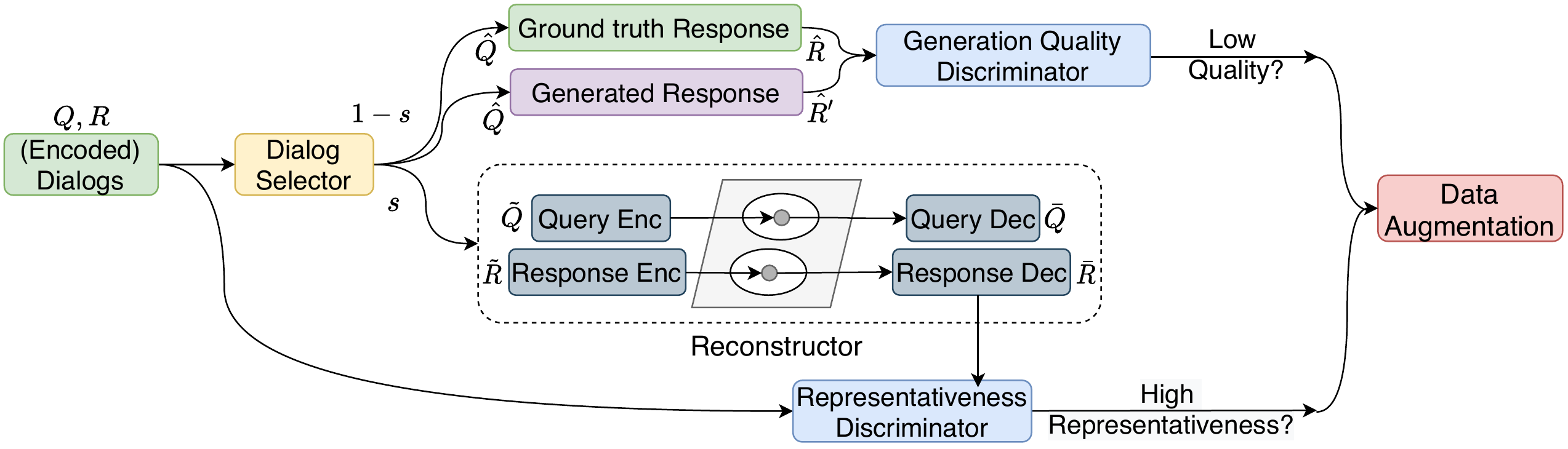}
		\caption{Major components of our approach. 
			The \textit{dialog selector} selects samples that will be examined by two discriminators.
			The \textit{generation quality discriminator} examines the generation quality of the selected cases, and the \textit{representativeness discriminator} examines the representativeness of the selected samples.
			The samples with low quality and high representativeness, \ie high $s$ score, will be selected for data augmentation.
			% 	To examine the generation quality of the selected cases, there is the generation uncertainty discriminator that discriminates between the ground-truth response and the generated response.
			% 	To examine the representativeness of the selected samples, a reconstructor first embeds selected samples and then reconstructs them. 
			% 	The representativeness discriminator is required to classify whether the input belongs to the original samples or the selected samples.
		}
		\label{fig:model}
	\end{figure*}

% 	\subsection{Dual Adversarial Framework}
	Overall, the \textit{Dialog Selector} assigns select weights to existing samples.
	To select the lowest quality and most representative cases, we propose two discriminators to assist this process, as shown in Figure~\ref{fig:model}.
	Firstly, a \textit{Generation Quality Discriminator} (GQD) discriminates between the ground-truth response and the generated response.
	The dialog selector will assign high weights to cases that cannot fool GQD.
	Secondly, to examine the representativeness of the selected samples, a reconstruction and a discrimination process are employed.
	The intuition is that if the selected cases can successfully reconstruct the original data, then the selected cases are representative.
	Concretely, a reconstructor first embeds the selected samples into the same latent space and then reconstructs them. 
	A \textit{Representativeness Discriminator} (RD) is then required to classify whether the input belongs to the original samples or the selected samples.
	Dialog selector will assign high weights to cases that fool RD.

	\subsection{Dialog Selector}
	\label{scorenet}
	
	We first employ a bi-directional recurrent neural network (Bi-RNN) to model the temporal interactions between words in the query and response, denoted by $Q=\{h^{q_i}_1,...,h^{q_i}_{m'}\}$ and $R=\{h^{r_i}_1,...,h^{r_i}_m\}$, respectively.
	$i$ denotes the sample index.
	% \begin{align}
	%     h^{q_i}_t &= \text{Bi-RNN}_q(e(q^i_t), h^{q_i}_{t-1}),\\
	%     h^{r_i}_t &= \text{Bi-RNN}_r(e(r^i_t), h^{r_i}_{t-1}), 
	% \end{align}
% 	Following~\cite{Tao2018GetTP}, we choose long short-term memory (LSTM) as the cell for Bi-RNN. 
	The final hidden state $h^{q_i}_{m'}$ and $h^{r_i}_{m}$ denotes the overall representation for the query and response.
	The dialog selector adopts a simple architecture, consisting of a 5-layer multi-layer perceptron (MLP) with Xavier initialization, to map the input feature to a score:
	\begin{align}
	\label{sigma}
	s^i=\sigma\left(\text{MLP}_a([h^{q_i}_{m'};h^{r_i}_m])\right),
	\end{align}
	where $[;]$ denotes the concatenation operation and $\sigma$ denotes the sigmoid function.
% 	The cases with high scores will be employed for data augmentation.
	
	In the next subsections, we will propose dual discriminators to assist the selection process.
	As preparation, the original representation $Q$ and $R$ are weighted using these scores, and we use $R$ to denote this process:
	\begin{align}
	\hat{R}^i=(1-s^i)R^i,
	\tilde{R}^i=s^iR^i.
	\end{align}
	$\hat{R}^i$ is employed for quality discrimination, and $\tilde{R}^i$ is used for representativeness discrimination.
	Note that we use $1-s^i$ and $s^i$ as the weights for quality and representativeness branches, respectively, to ensure the optimization of these two terms in the same direction.
	Notations for $\hat{Q}^i$ and $\tilde{Q}^i$ are similar.
	% Notation for response is similar.
	
	To prevent the selector from assigning equal importance to all data points, we employ a length regularizer loss $ \mathcal{L}_{\mathrm{LR}}$ to limit the number of selected elements, and use a determinantal point process (DPP) loss $\mathcal{L}_{\text {dpp}}$ \cite{szegedy2015going} to ensure the diversity of selected data points:  
	\begin{align}
	    \mathcal{L}_{\mathrm{LR}}=\left\|\sigma-\frac{1}{N} \textstyle \sum_{i=1}^{N} s^i\right\|_{2},\mathcal{L}_{\mathrm {dpp}}=-\log (P(\boldsymbol{s})).
	\end{align}
For $\mathcal{L}_{\mathrm{LR}}$, $\sigma$ represents the percentage of cases for subset selection.
For $\mathcal{L}_{\mathrm{dpp}}$, $P(\boldsymbol{s})$ is a probability that DPP assigns to the select $\boldsymbol{s}$.
We compute $P(\boldsymbol{s} ; L)=\frac{\operatorname{det}(L(\boldsymbol{s}))}{\operatorname{det}(L+I)}$, where $L$ is an $N \times N$ similarity matrix between every case, $I$ is an identity matrix, and $L(\boldsymbol{s})$ is a smaller square matrix, cut down from $L$ given $\boldsymbol{s}$. 
For $i$-th case and $j$-th case, the pairwise similarity values are defined as $L_{i,j}=s^i s^j [h^{q_i}_{m'};h^{r_i}_m] [h^{q_j}_{m'};h^{r_j}_m]$.

	\subsection{Generation Quality Discriminator}
	
% 	\begin{figure}
% 		\centering
% 		\includegraphics[width=0.8\linewidth]{augmentation-discriminator.pdf}
% 		\caption{Illustration of the generation quality discriminator.
% 		}
% 		\label{fig:dis}
% 	\end{figure}

	% Note that our framework is flexible since it does not have restrictions on the response generation model type.

	Generation quality discriminator (GQD) aims to evaluate whether the generated responses are feasible for a given query.
	We achieve this by measuring the matching degree between query-response pairs in an adversarial fashion.
	The weighted ground truth query-response pair is treated as a positive case, while the query with the generated response pair is the negative case.
	Concretely, for the positive pair, we concatenate the weighted ground-truth response $\hat{R}^i$ with the weighted query $\hat{Q}^i$.
	Then, a fully-connected neural network with a sigmoid activation function is utilized to flatten the feature matrix, resulting in the final matching score $m^i_{g} \in(0,1)$. 
	The matching score $m^i_f$ between the negative instance $\hat{R}^{'i}$ and query $\hat{Q}^i$ is also calculated as the above-mentioned procedure, except that we have a dimension transformation on the generated response to align it with $\hat{Q}^i$. 
	Note that our framework does not rely on specific response generation models, and in our case, we employ LSTM-based RNN as the generator.
	
	In the paradigm of GAN, the training objective of GQD is to maximize the matching score of positive instances and minimize the negative ones,	while the dialog selector is optimized by aiming to maximize the matching score the generated response:
	\begin{align}
	\label{gan}
	\mathcal{L}_{D}&=-\textstyle \sum_{i=1}^{N}\left(\log(1-m^i_f)+\log(m^i_g)\right),\\
	\mathcal{L}_{G}&=-\textstyle \sum_{i=1}^N \left(\log(m^i_f)\right).
	\end{align}
	The dialog selector will learn to assign high weights, \ie $1-s^i$, to samples that are difficult for GQD to identify, which leads to a low $s^i$ score.
	In other words, the cases that obtain high $s^i$ scores have low quality.

	\subsection{Reconstructor}
	In the next two subsections, we introduce our representativeness selection process, where the samples that can be used to reconstruct the original dataset are selected.
% 	Our intuition for selecting representative samples is that, if the selected samples can be used to reconstruct the original dataset, then the selected samples can be used as surrogates for the full dataset.
	This is inspired by \citet{Mahasseni2017UnsupervisedVS}, where they address the problem of finding representative images in a video.
	Their key idea is to find a subset of images that can be used to reconstruct the whole video.
	In this work, we extend this ideology from video level to dataset level to find representation cases instead of images.
	Since dialog data is paired, we use the query to illustrate this process.

	Our reconstructor takes the form of VAE, which is commonly used to effectively learn feature representations \cite{Zhao2017LearningDD}. 
	VAE defines a posterior distribution over the observed data, given an unobserved latent variable.
	Overall, VAE consists of an encoder and a decoder.
	The encoder maps the weighted query $\tilde{Q}^i$ to a latent space $e$, and the decoder reconstructs the query from $e$.
	
	Concretely, the encoder computes posterior distributions $q_\theta(e|\tilde{Q}^i)$, where the latent representations $e$ is sampled.
	The reconstruction process can be formulated as $p_\theta(\tilde{Q}^i|e)$, representing the probability of generating input $Q^i$ conditioned on $e$. 
	Herein $\theta$ represents the parameters of the above encoders and reconstruction decoder.
	Because of the intractable integral of the marginal likelihood $p_\theta(\tilde{Q}^i)$, the posterior $q_\theta(e|\tilde{Q}^i)$ is simulated by variational approximation $q_\phi(e|\tilde{Q}^i)$, where $\phi$ is the parameters for $q$.
	When learning the VAE, the objective is to maximize the variational lower bound of $\log p_\theta(\tilde{Q}^i)$:
	\begin{align*}
	\mathcal{L}_{VAE}^q
	={\rm KL}(q_\phi(e|\tilde{Q}^i)\|p_\theta(e))	-{\mathbb{E}}_{q_\phi(e|\tilde{Q}^i)}[{\rm log}p_\theta(\tilde{Q}^i|e)],
	\end{align*}
	where the $\text{KL}$ denotes KL-divergence, the regularization for encouraging the approximated posterior $q_\phi(e|\tilde{Q}^i)$  to be close to the prior $p_\theta(e)$, \ie standard Gaussian distribution.
	${\mathbb{E}}[\cdot]$ is the reconstruction loss conditioned on the approximation posterior $q_\phi(e|\tilde{Q}^i)$.
	
		We denote the reconstructed query as $\bar{Q}^i$, and reconstructed response as $\bar{R}^i$.
	
% 	Similarly, we denote the reconstructed response as $\bar{R}^i$.
	
	\subsection{Representativeness Discriminator}
	The discrimination processes for query and response are similar, and we use the query to illustrate this process.
	Representativeness discriminator (RD) takes $\bar{Q}^i$ and $\tilde{Q}^i$ as input and aims to classify them into two distinct classes (\ie selected or original).
	RD adopts the same architecture in GQD except that it does not have the dimension transformation.
	We omit the details here due to limited space.
	RD aims to maximize the correct matching result, while the dialog selector aims to select cases that can fool RD.
	If RD cannot distinguish the selected cases from the original one, the dialogs with high $s^i$ scores are seen to have good representativeness to the dataset.
	Hence, the dialog selector will learn to assign high $s^i$ score to the representative case to fool RD.
	
%	\subsection{Training strategy}
%	Our SDA involves two adversarial games between the dialog selector and dual discriminators, which iteratively optimize the dialog selector for accurate data selection. 
%	For GQD, if a low-quality generated case cannot it, the higher score should be maintained. 
%	Simultaneously, the adversarial learning between VAE and RD encourages the dialog selector to assign larger scores to the samples representative to the dataset.
%	Therefore, the importance scores yielded by our dialog selector can simultaneously evaluate the uncertainty and representativeness of data points. 
%	During the selective data augmentation process, the samples with top largest scores ranked by the dialog selector are selected from the training dataset to augment.

	\section{Experiment}

	\subsection{Experiment Setup}
	\label{data}
	\textbf{Datasets.} Following \citet{Cai2020DataMT}, we conduct experiments on two English conversation datasets: 
	(1) \textit{DailyDialog}~\cite{Li2017DailyDialogAM}, a collection of real-world dialogues widely used in open-domain dialogue generation.
	This is a multi-turn dataset, and we treat each turn as a training pair in this work.
	The overlapping pairs are removed from the dataset.
	(2) \textit{OpenSubtitles}~\cite{Lison2016OpenSubtitles2016EL}, a group of human-human conversations converted from movie transcripts.
	We split the DailyDialog  dataset to 54,889/6,005/5,700, and OpenSubtitles to 64,000/8,000/8,000.

% 	\subsection{Implementation Details}
	
	\textbf{Implementation Details.} (1) \textit{Hyperparameter setting}: We implement our models in TensorFlow on an NVIDIA GTX 1080 Ti GPU. 
	We truncate the input dialog to 20 words, the minimum decoding step is 10, and the maximum step is 30.
	The default $\sigma$ in Equation~\ref{sigma} is set to 0.6 except in the augmentation percentage analysis.
% 	The word embedding dimension is set to 128 and the number of hidden units is 256.
	The batch size is set to 16, and we limit the vocabulary size to 50K.
% 	We use Adagrad optimizer as our optimizing algorithm with 0.15 learning rate.
% 	During the inference stage, the checkpoint with the smallest validation loss is chosen and the beam-search size is set to 4 for all methods.
	(2) \textit{Optimization techniques:}
% 	Firstly, we use the Bag-Of-Words (BOW) loss along with KL annealing of 10,000 batches to address the vanishing latent variable problem \cite{Bowman2016GeneratingSF}.
	We employ a set of techniques to deal with the posterior collapsed problem in VAE \cite{Bowman2016GeneratingSF} including Bag-Of-Words (BOW) and KL annealing.
	We increase the kl loss coefficient by 0.5 every 10,000 batches.
	Readers can refer to work by \cite{Zhao2017LearningDD} for details.
	For the GANs in our framework, we train the discriminator for one step every five steps for the generator, since it is it would be harder for generation than classification.
	The generators and discriminators are adversarially trained until GQD cannot discriminate between ground-truth and generated responses and RD is not able to distinguish between the summary and original datasets.
	The framework comes to convergence in less than an hour.
    (3) \textit{Augmentation details: }
	We select 60\% cases with the highest scores for augmentation if not specified, based on the experiment result on the validation dataset.
	For the selected cases, we employ the back-translation technique~\cite{Sennrich2016ImprovingNM} to augment them by ten times following \citet{Li2019InsufficientDC}.
	We choose the back-translations since it provides more diverse augmented text with different structures while preserving the meaning of the original text \cite{einolghozati2019improving,chen2021empirical}.
	% We use translation systems provided by fairseq with two auxiliary languages: Russian and German, and keep the top 5 beams.
	% Note that other data augmentation techniques can also be applied in our framework, and we leave it as future work.
 Our evaluation metrics include distinctness~\cite{Li2016ADO}, BLEU~\cite{papineni2002bleu}, and embedding metrics \cite{gu2018dialogwae}.

	\textbf{Baselines.}
	We compare our model on following classic generation structure:
	\noindent(1) \textbf{SEQ2SEQ} \cite{Bahdanau2015NeuralMT}: a sequence-to-sequence model with attention mechanisms.
	\noindent(2) \textbf{CVAE} \cite{Zhao2017LearningDD}:  a latent variable model using conditional variational auto-encoder, trained with KL-annealing and a BOW loss.
	\noindent(3) \textbf{Transformer} \cite{Vaswani2017AttentionIA}:  an encoder-decoder architecture relying solely on the attention mechanisms.
	\noindent(4) \textbf{GPT-2} \cite{radford2019language}: a  large-scale pre-trained language model, which is finetuned by the full training dataset.
	We also compare our approach with native augmentation, previous data augmentation, or instance weighting methods:
	\noindent(1) \textbf{Random}: we randomly select 60\% data for augmentation, to compare with our selective augmentation method.
    Comparisons with different augmentation percentages can be found in the discussion section.
	\noindent(2) \textbf{Calibration}~\cite{Shang2018LearningTC}: a calibration network measures the quality of data samples and enables weighted training for dialogue generation.
	\noindent(3) \textbf{CVAE-GAN}~\cite{Li2019InsufficientDC}:  a model that combines CVAE and GAN for augmentation.
	\noindent(4) \textbf{Manipulation}~\cite{Cai2020DataMT}: it augments all the cases in the training process and reweights them.

		\begin{table*}[tb]
		\centering
		\resizebox{0.8\textwidth}{!}{	\begin{tabular}{c|l|ccc|cccc|ccc}
				\toprule
				& Models & Dist-1 & Dist-2 & Dist-3  & BLEU-1& BLEU-2& BLEU-3& BLEU-4 & Avg & Ext & Gre \\
				\midrule
				\multirow{8}{*}{(a)} & SEQ2SEQ & 1.36 & 5.98 & 11.19   &12.85 &2.26 &1.12 &0.93 & 76.56 & 42.98 & 62.24\\ 
				&SEQ2SEQ (${\bigstar}$) & \textbf{1.63} & \textbf{7.24} & \textbf{15.01}  & \textbf{14.65} &\textbf{2.73} &\textbf{1.27} &0.81& \textbf{77.83} & \textbf{43.60} & \textbf{63.56}   \\ \cline{2-12} 
				&CVAE & 1.88 & 6.50& 12.08  & 11.38 &2.08 &1.10 &0.75 & 74.18 & 40.71 & 62.49 \\  
				&CVAE (${\bigstar}$)    &  \textbf{3.26 }&\textbf{ 13.57} & \textbf{22.41 }  &\textbf{13.49} &\textbf{3.31} &\textbf{1.30} &\textbf{0.92} & \textbf{75.72} & \textbf{42.30 }& \textbf{63.68} \\ \cline{2-12} 
				&Transformer & 1.19 & 6.21 & 15.13  &  13.29 &2.28 &1.13 &0.76 & 76.38 & 44.07 & 62.73 \\ 
				&Transformer (${\bigstar}$) & \textbf{2.92} & \textbf{15.20 }&  \textbf{23.70 }& \textbf{14.17 }&\textbf{2.52} & \textbf{1.33}&\textbf{0.85} & \textbf{77.46} & \textbf{45.69} &  \textbf{64.20}   \\
				\cline{2-12} 
				&GPT-2 & 2.16 & 7.44 & 16.15  &  15.27 &2.84 &1.66 &0.78 & 78.27 & 45.39 & 64.17 \\ 
				&GPT-2 (${\bigstar}$)  &\textbf{2.57} &  \textbf{9.06} & \textbf{19.54}&\textbf{16.29} & \textbf{3.23}&1.54& \textbf{0.81}&\textbf{79.15} & \textbf{46.23 }&  \textbf{64.65} \\
				\bottomrule
				\bottomrule
				\multirow{8}{*}{(b)} & SEQ2SEQ 
				& 1.37& 2.22& 6.62& 10.03&1.57 & 1.01& 0.84 & 61.88 & 45.34 & 50.45 \\  
				&SEQ2SEQ (${\bigstar}$)     &  \textbf{1.56} & \textbf{3.94} & 5.83 &\textbf{10.78 }&\textbf{2.00} &\textbf{1.29 }&\textbf{0.97} & \textbf{62.36} & \textbf{46.24 }& \textbf{51.14 }  \\ \cline{2-12} 
				&CVAE 
				& 0.70 & 2.22 & 5.92 & 9.90 &1.87 &1.07 &0.92 & 65.37 & 49.60 & 53.63\\ 
				&CVAE (${\bigstar}$)  
				&\textbf{1.88} & \textbf{4.29} & \textbf{9.40}  & \textbf{11.15} &\textbf{2.09} & \textbf{1.18}&\textbf{0.93}& \textbf{67.64} & \textbf{50.74} & \textbf{54.72} \\ \cline{2-12} 
				&Transformer &1.57 & 3.28 & 6.39   &8.76 &2.35 &1.21 &0.87 & 66.91 & 44.40 & 54.18 \\
				&Transformer (${\bigstar}$) & \textbf{2.92} & \textbf{7.38} & \textbf{10.14} &\textbf{10.35} & \textbf{2.60}&\textbf{1.49} &\textbf{0.91} & \textbf{68.04} & \textbf{45.99} & \textbf{54.96}\\ 
				\cline{2-12} 
				&GPT-2 &3.12 & 4.32 & 7.29   &10.97 &3.30 &2.15 &1.15 & 67.28 & 48.60 & 55.07 \\
				&GPT-2 (${\bigstar}$) & \textbf{3.51} &\textbf{5.36} & \textbf{8.59 }&\textbf{11.63}& \textbf{3.56}&\textbf{2.45} &\textbf{1.17} & \textbf{68.37} & \textbf{49.18} & \textbf{55.50}\\ 
				\bottomrule
		\end{tabular}}
		\caption{Automatic evaluation results (\%) on (a) DailyDialog and (b) OpenSubtitles. 
			``${\bigstar}$'' denotes that the model is trained using our proposed framework.
			{The metrics Average, Extrema, and Greedy are abbreviated as Avg, Ext, and Gre, respectively.}
			Numbers in bold mean that the improvement to the best baseline is statistically significant (a two-tailed paired t-test with p-value \textless 0.01).
		}
		\label{tbl:main_res}
	\end{table*}

		\begin{table*}[tb]
		\centering
		\resizebox{1\textwidth}{!}{
			\begin{tabular}{l|ccc|cccc|ccc}
				\toprule
				& Dist-1 & Dist-2 & Dist-3 & BLEU-1 & BLEU-2 & BLEU-3 & BLEU-4 & Avg & Ext & Gre \\
				\midrule
				Full model
				&\textbf{1.63} & \textbf{7.24} & \textbf{15.01} & \textbf{14.65}&\textbf{2.73 }&\textbf{1.27} &0.81& \textbf{77.83} & \textbf{43.60} & \textbf{63.56}  \\ 
				\quad \textit{w/o selective augmentation} 
				& 1.42 \down & 5.03\down & 12.98 \down &13.31&2.30 &1.15 & 0.68& 76.26  \down & 42.45\down& 62.54 \\ 
				\quad \textit{w/o quality discriminator} 
				& 1.59 \down & 5.55\down & 13.73 \down &13.26&2.55  &1.18 & 0.96 &77.32 \down & 42.78 \down & 62.67\down \\ 
				\quad \textit{w/o representativeness discriminator} 
				& 1.55 \down & 6.23\down & 13.21 \down &14.02\down &2.66 & 1.25 & 0.73 & 76.71 & 42.83 \down  & 63.92 \down \\ 
				\bottomrule
			\end{tabular}
		}
		\caption{Ablation test of our model (\%) on DailyDialog, which is instantiated on the naive SEQ2SEQ.
	Numbers in bold mean that the improvements to the ablation models are statistically significant. 
			}
		\label{tbl:ablation_model}
	\end{table*}

	\begin{table*}[tb]
		\centering
		\resizebox{1\textwidth}{!}{
			\centering
			\begin{tabular}{c|l|ccc|cccc|ccc}
				\toprule
				& Models & Dist-1 & Dist-2 & Dist-3 & BLEU-1 & BLEU-2 & BLEU-3&BLEU-4 & Avg & Ext & Gre  \\ 
				\midrule
				\multirow{4}{*}{(a)} 
				& Random & 1.38& 5.24& 11.70 & 13.25 &2.17 & 1.13 & 0.73 & 77.02 & 43.12 & 62.68  \\
				&				{Calibration}~\cite{Shang2018LearningTC} & 1.53& 5.96& 11.77 &13.09 & 2.28 & 1.02& 0.75 &77.15 & 42.94 &62.77  \\ 
				& {CVAE-GAN}~\cite{Li2019InsufficientDC} & 1.54& 5.63& 13.50 & 14.00 &2.59& 1.24&0.98 & 77.21 &43.19 & 62.96  \\ 
				& {Manipulation}~\cite{Cai2020DataMT} &1.58& 6.42& 14.52  & 14.26 &2.87& 1.16&0.95 & 77.53 &43.32 & 63.12  \\ 
				& SDA & \textbf{1.63} & \textbf{7.24} & \textbf{15.01}  & \textbf{14.65} & \textbf{2.92} & \textbf{1.27} &0.81 & \textbf{77.83} & \textbf{43.60} & \textbf{63.56 } \\  
				\bottomrule
				\bottomrule
				\multirow{4}{*}{(b)} & 				{Random} & 1.40 &  2.46 & 5.76 &  10.05 & 1.22 & 1.03 & 0.93 &  61.97 & 45.51 & 50.72 \\ & {Calibration}~\cite{Shang2018LearningTC} &1.43&  2.58 & 5.82 &  10.20&1.23&1.08 &0.68 &  62.03&45.57 & 50.83  \\  
				& {CVAE-GAN}~\cite{Li2019InsufficientDC} &1.49& 2.83& 5.07 & 10.26& 1.28&1.17 & 0.87&62.28 &45.74 & 50.76  \\  
				& {Manipulation}~\cite{Cai2020DataMT} &1.41& 3.40& 5.93 & 10.37 & 1.58& 1.24& 0.94& 62.29 &46.00 &50.22   \\  
				& SDA  &\textbf{1.56} & \textbf{3.94} & 5.83 &\textbf{10.78} &\textbf{2.00} &\textbf{1.29} &\textbf{0.97} & \textbf{62.36} & \textbf{46.24} &\textbf{ 51.14}   \\ 
				\bottomrule
			\end{tabular}
		}
			\caption{Performance (\%) of our approach instantiated on the naive SEQ2SEQ and the baseline approaches on (a) DailyDialog and (b) OpenSubtitles.
		Numbers in bold mean that the improvement to the best baseline is statistically significant. 
		}
		\label{tbl:compare_res}
	\end{table*}

	\subsection{Main Results}
	\label{main}

	\textbf{Automatic evaluation.}
	We instantiate our framework on a number of classic dialog generation models including SEQ2SEQ, CVAE, Transformer, and GPT-2.
	The automatic evaluation results are shown in Table~\ref{tbl:main_res}.
	It can be seen that our model outperforms vanilla baselines on almost all automatic metrics. 
	The improvements are consistent across both datasets, demonstrating the superiority and general applicability of our framework.
	
	In addition, we compare our model with the existing augmentation methods.
	We select SEQ2SEQ as the response generation model following since all compared models are constructed on this classic baseline~\cite{Cai2020DataMT}.
	Not surprisingly, as shown in Table~\ref{tbl:compare_res}, our framework outperforms most of the baseline methods.
	Concretely, SDA outperforms Random baseline in all metrics, demonstrating that selection is necessary to improve the performance of data augmentation.
	 CVAE-GAN augments each case in the training dataset, and Manipulation augments every case in each training step, while our model only augments 60\% data and achieves better performance.
	This demonstrates that selective data augmentation is more effective and efficient, outperforming data augmentation methods that require generating more augmented cases. 
	The statistical significance of observed differences between the performance of two runs is tested using a two-tailed paired t-test for $\alpha = 0.01$.
	
\begin{table}
		\small
		\centering{\begin{tabular}{ccc}
				\toprule
				Model& Readability&Informativeness\\
				\midrule
				Calibration&1.63& 1.68\\
				CVAE-GAN& 1.85&1.81\\
				Manipulation& 1.91&2.07\\
				SDA& \textbf{2.01}&\textbf{2.12}\\
				\bottomrule
		\end{tabular}}
		\caption{Human evaluation on two aspects: Readability and informativeness. }
		\label{tab:human_evaluation}
		% \vspace{-4mm}
	\end{table}
	
		\begin{table}[htb]
			\renewcommand\arraystretch{0.8}
			\centering
			\resizebox{0.9\columnwidth}{!}{
				\begin{tabular}{c|l}
					\toprule
					\multicolumn{2}{l}{\begin{tabular}[c]{@{}l@{}}-I got a ticket yesterday.\end{tabular}} \\
					\midrule
					\multirow{1}{*}{ground-truth}   & Really? What did you get one for?    \\
					\multirow{1}{*}{CVAE-GAN}     & Is that right?? Is that possible ? \\ 
					\multirow{1}{*}{Manipulation}   & 88 yuan, please.  \\
					\multirow{1}{*}{SDA}   & Really? How much is it?  \\
					\bottomrule
					\toprule
					\multicolumn{2}{l}{\begin{tabular}[c]{@{}l@{}}- What do you mean?  You have a lover? .\end{tabular}} \\
					\midrule
					\multirow{1}{*}{ground-truth}   & A fiance. \\
					\multirow{1}{*}{CVAE-GAN}     & You've had a lot of your own! lover! \\ 
					\multirow{1}{*}{Manipulation}   & No, I'm serious.    \\
					\multirow{1}{*}{SDA}   & Yeah, she's so different, she is the sun! \\
					\bottomrule
				\end{tabular}
			}
			\caption{Responses generated by baselines and our model. The  top case  is selected  from DailyDialog, and the bottom case is from OpenSubtitles.}
			\label{tab:casestudy}
		\end{table}

\begin{figure*}
  \centering
  \includegraphics[scale=0.27]{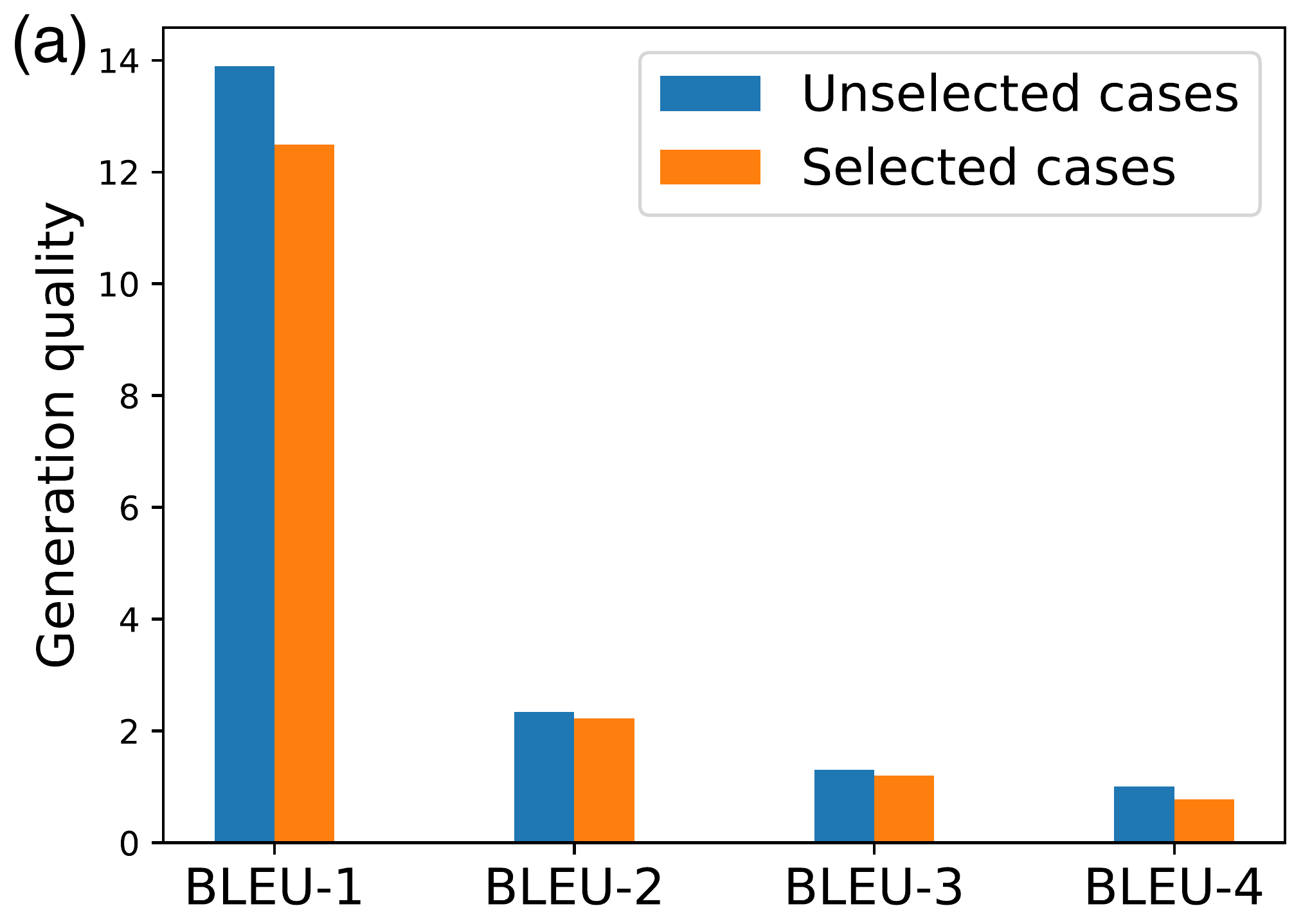}
  \includegraphics[scale=0.27]{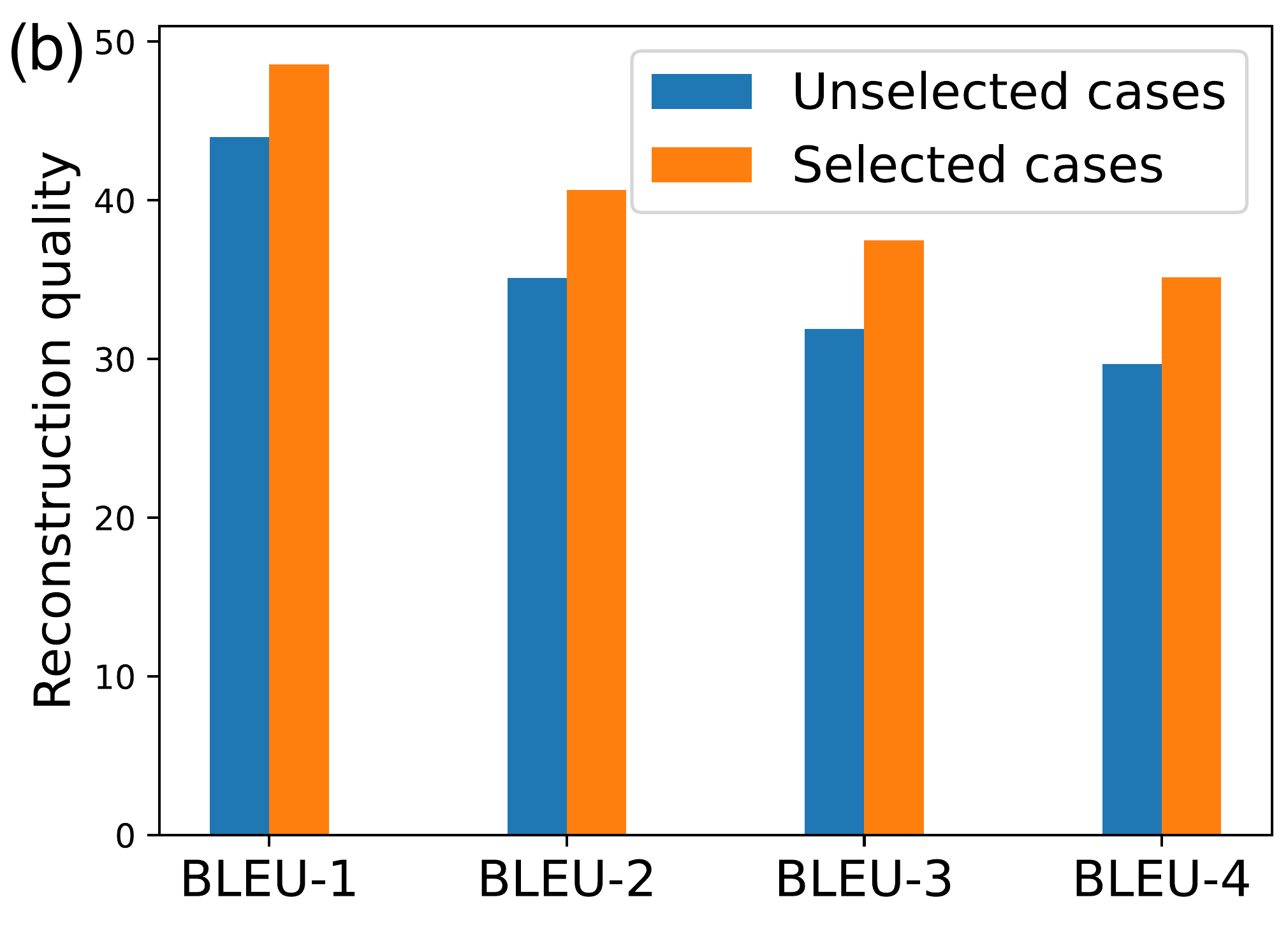}
  \includegraphics[scale=0.27]{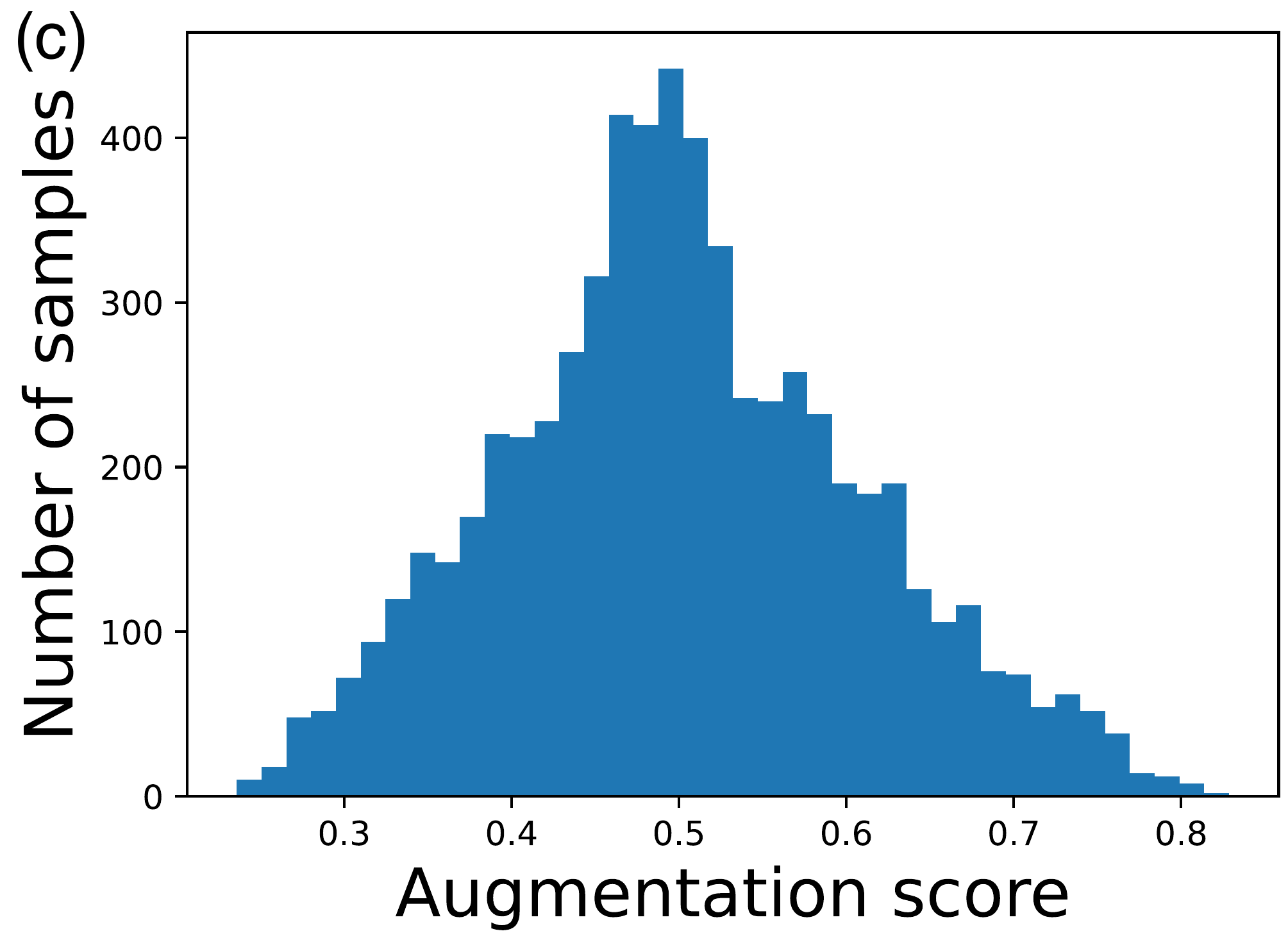}
  \caption{(a) Generation quality of selected and unselected cases. (b) Reconstruction performance of selected and unselected cases. (c) Histogram of different augmentation frequencies. }
  \label{contras}
\end{figure*}

    \textbf{Human Evaluation.}
    \label{manual}
	We also employ a human evaluation on Amazon Mechanical Turk. 
	For better annotation quality, we employ three annotators and require the annotators to have at least 99\% approval rate with at least 1,000 approved HITs.
	These annotators are hired to evaluate the quality of generated responses on DailyDialog dataset, where the evaluation is conducted in a double-blind fashion.
	Totally, 200 randomly sampled responses generated by each model are rated by each annotator with two different aspects, \ie \textit{readability} and \textit{informativeness}. 
	Criteria are scored from 1 to 3, \ie bad, normal, and good.
	The results of the human evaluation are listed in Table~\ref{tab:human_evaluation}. 
	Our model significantly outperforms most of the baselines in terms of all the metrics. 
	Particularly, our model increases informativeness by approximately 2.4\% over Manipulation. 
	The kappa statistics is 0.42 and 0.45 for readability and informativeness, respectively, which indicates moderate agreement between annotators.
	We also show a representative case from DailyDialog in Table~\ref{tab:casestudy},
	It can be seen that our model can generate a more diverse and interesting response that describes in detail how it feels to have a lover.

	\subsection{Discussions}
	
	\textbf{Ablation Study.}
	We also list the results of the ablation study in Table~\ref{tbl:ablation_model}, aiming to investigate the influence of different modules in our proposed model.
	It can be seen that the performance of all metrics drops if we direct augment all the training data without selection.
	This demonstrates that selection is important for augmentation.
	We also find that the Dist-1 score drops by 2.45\% and 4.91\% after GQD and the RD are removed, respectively.
	This indicates that the jointly selected cases from the quality and representativeness aspects help generate more diverse and accurate responses.

\begin{figure*}[htb]
  \centering
  \includegraphics[scale=0.28]{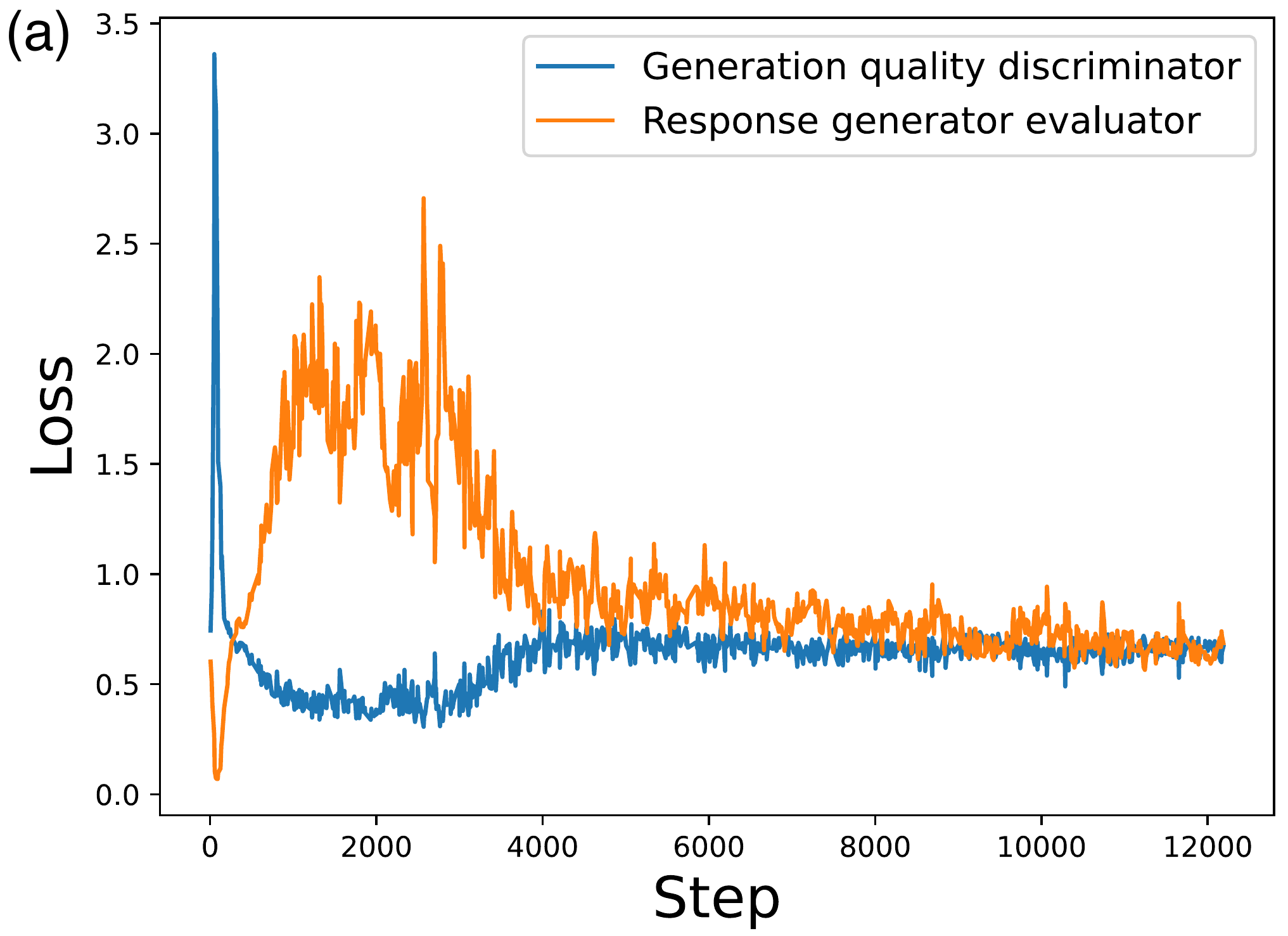}
%   \hspace{1in}
  \includegraphics[scale=0.28]{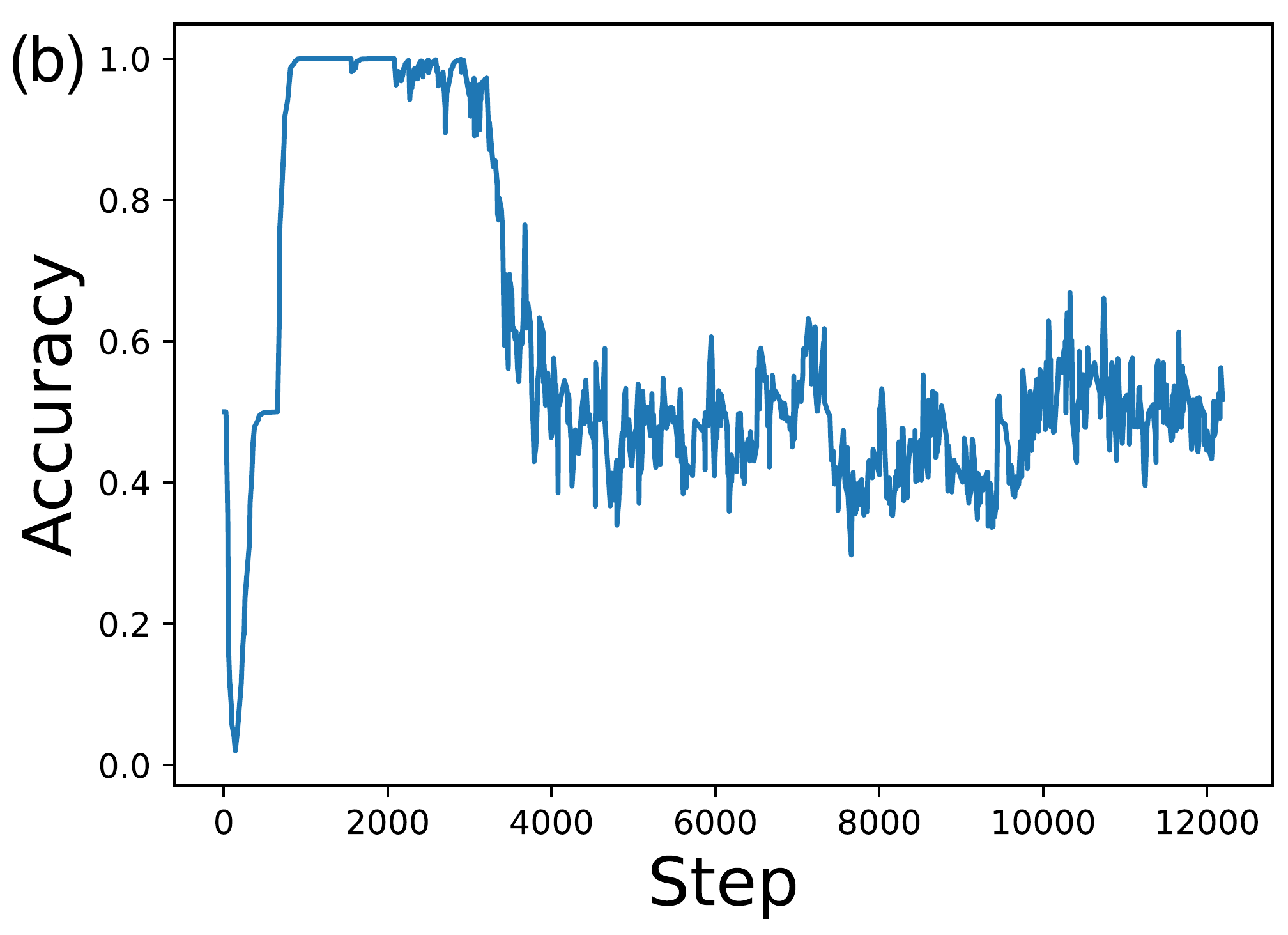}
%   \hspace{1in}
  \includegraphics[scale=0.22]{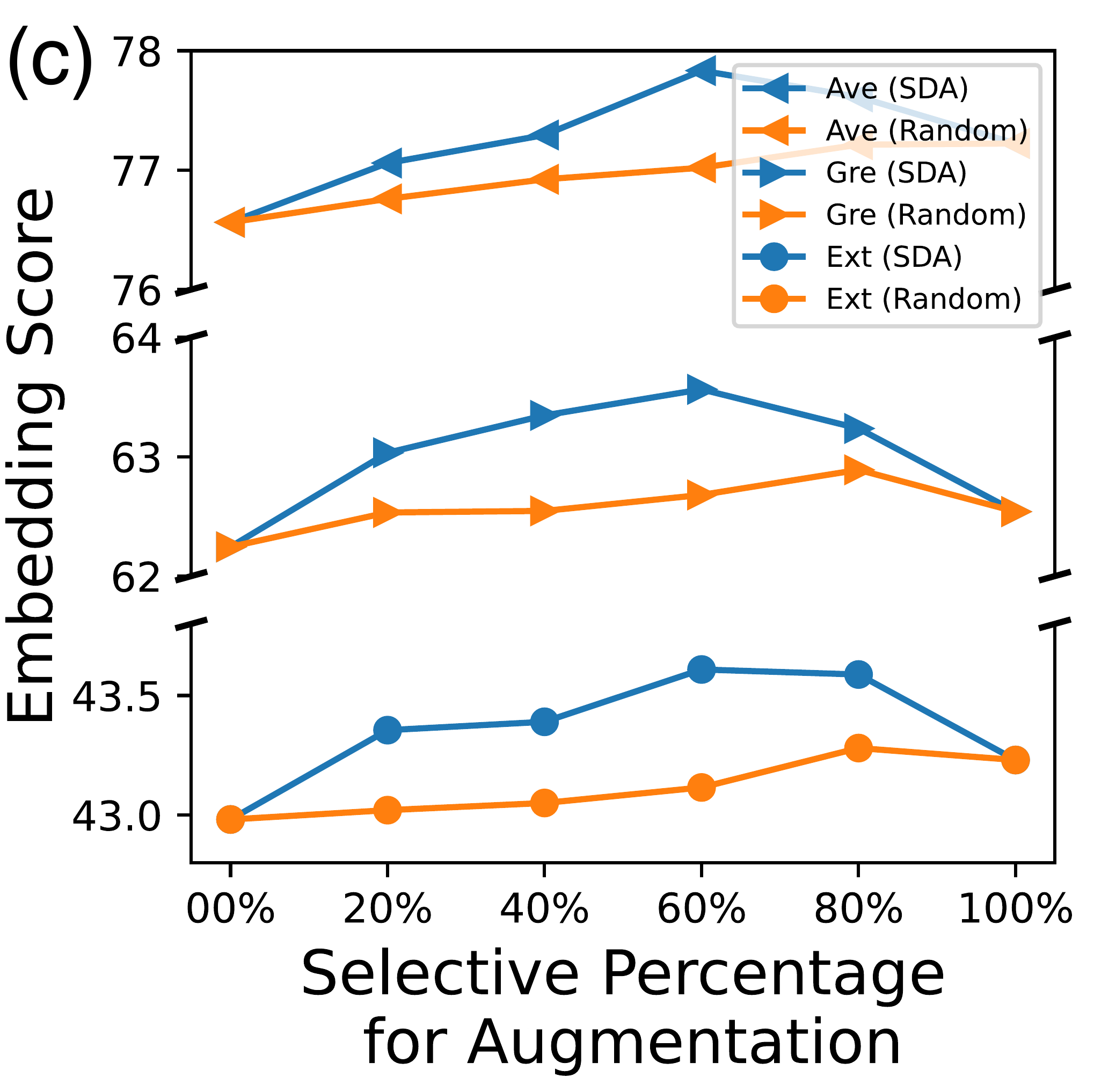}
  \caption{ (a) Loss curve of the quality discriminator and generator evaluator. (b) Accuracy curve of the quality discriminator. (c) Relationship between the selective percentage for augmentation and the embedding scores. Blue denotes our model, and orange denotes the Random model. }
  \label{f2}
\end{figure*}

\begin{table*}[htb]
\centering
\small
\begin{tabular}{c|l|cccc|ccc}
\toprule
&Models      & BLEU-1 & BLEU-2 & BLEU-3 & BLEU-4                 
& Extrema         & Average         & Greedy          \\ \midrule

\multirow{3}{*}{RocStories}
& CVAE & 25.81 & 9.69 & 3.60 & 1.48 & 51.35 & 56.32 & 60.11 \\
& Seq2Seq & 23.24 & 8.96 & 3.40 & 1.50 & 51.33 & 56.49 & 60.07 \\
& Transformer & 25.52 & 5.96 & 3.54 & 1.45 & 51.29 & 56.37 & 60.06 \\
& GPT-2 & 30.21& 11.08 & 3.64 & 1.53 & 51.72 & 58.49 & 60.35 \\
\cline{2-9}
& GPT-2(${\bigstar}$) & \textbf{30.96} & \textbf{11.46} & \textbf{3.84} &1.52 & \textbf{52.27} & \textbf{58.95}& \textbf{61.20} \\
\bottomrule
\end{tabular}
\caption{Automatic evaluation results on RocStories for storytelling. 
Numbers in bold mean that the improvement to the best baseline is statistically significant (t-test with p-value \textless 0.01).}
\label{tab:automatic_results}
\end{table*}

	\textbf{Analysis of Selected Samples.}
	In this subsection, we examine whether the model successfully selects the lowest quality and most representative cases for augmentation.
	We calculate the BLEU scores of selected and unselected cases in the response generation task and response reconstruction task.
	From Figure \ref{contras}(a) and Figure~\ref{contras}(b) we can see that the selected cases have lower BLEU scores in terms of the generation quality and higher scores in the reconstruction task.
	This demonstrates that the model needs to be polished to generate better responses for the selected cases.
	In the meantime, the selected data itself is not noise data and represents the overall data distribution.
	To further glean the insights regarding which samples are favored by the augmentation model, we also list examples with different augmentation scores in Figure~\ref{contras}(c).
	We notice that samples frequently augmented by SDA are more reliable and meaningful context, where the response is closely related to the query and leads to a new topic.
	 While for the dialog pairs seld augmented, they contain universal and safe content such as ``I don't know'' or ``I'd forgotten about it''.

	\textbf{Visualization of Dual Training.}
    To visualize the select process, we draw the loss curve of the generation quality discriminator ($\mathcal{L}_D$ in Equation~\ref{gan}) and the response generator evaluator ($\mathcal{L}_G$ in Equation~\ref{gan}) in Figure~\ref{f2}(a), and show the accuracy of GQD in Figure~\ref{f2}(b).
    When the training begins, the loss of the GQD and RGE fluctuates from time to time, as well as the accuracy curve, which verify the adversarial training. 
    After several steps, the training converges, and the accuracy of RD stays around 50\%, which means GQD cannot distinguish between the generated response and the ground truth one.
    In other words, the model successfully assigns low $s$ scores, \ie high (1-$s$), to the cases with high-quality generated responses so that GQD cannot perform better than a random guess. 
    The accuracy curve of the RD is similar to that of GQD, which proves that our model assigns high $s$ scores to the most representative cases so that RD cannot distinguish between the reconstructed and the original cases.

	\textbf{Impact of Augmented Data Scale.}
	For previous experiments, the percentage of cases for augmentation is set to 60\%.
	In this subsection, we change this percentage to study what is the influence of scale for augmentation and whether selective augmentation is still beneficial under different selection percentages.
	We also select the random baseline model for better comparison, where the cases for augmentation are randomly sampled.
	The result on the DailyDialog test dataset is shown in Figure~\ref{f2}(c).
	For Random baseline, its performance generally improves with the augmentation percentage. 
	This result shows that random augmentation will benefit the dialog generation task, and the more cases are augmented, the better performance will be obtained.
	However, this is not true for selective augmentation.
	It can be seen that to begin with, the embedding scores of SDA increase fast with the selective percentage for augmentation.
	After the percentage reaches 60\%, the growth stops, and when the percentage increases from 80\% to 100\%, there is even a drop in the performance.
	Similar performance is also observed on the OpenSubtitles dataset.
	This demonstrates that it only benefits the model if we select the proper cases in the dataset for augmentation, otherwise, augmenting some cases brings harm to the model.
	
	\textbf{Universality of our framework.}
	In addition, we test the generalization ability of our framework on the story generation task.
	RocStories dataset \cite{mostafazadeh2016corpus} consists of 98,163 high-quality hand-crafted stories, which capture causal and temporal commonsense relations of daily events.
	 Each story paragraph contains 5 sentences with an average of 43 words.
	 Following the previous work \cite{yu2021content}, we split the dataset into 8:1:1 for training, validation, and test, and use BLEU as the evaluation metric.
    As can be seen from Table \ref{tab:automatic_results}, equipped with augmentation data, our method outperforms \texttt{GPT-2}by 2.4\%, 3.4\%, and 1.4\% on RocStories in terms of BLEU-1, BLEU-2, and Greedy, respectively, which proves the superiority of our model.
    This experiment also demonstrates that our framework does not rely on a specific task, and can be extended to various text generation scenarios.

	\section{Conclusion and Broader Impacts}
	\label{conclusion}
	In this paper, we propose a selective data augmentation framework to improve the performance of dialogue models.  
	We propose a dual adversarial network to select data for augmentation from the quality and representativeness aspects. 
	One is to examine whether the case is of low generation quality, and the other one is whether the case is representative of the dataset.
	Experiments conducted on three public datasets demonstrate the effectiveness of our framework.
	In the future, we would like to explore the effectiveness of selective data augmentation on more generation tasks. 
	
% 	This work has the potential positive impact on an intelligent and engaging dialogue system.
% At the same time, this work may have some negative consequences on social interaction.
% Besides, if the training corpus includes malicious and vulgar information, it will bring bad information feedback to users.
% Therefore, we should be cautious of these advantages and disadvantages.

\section*{Acknowledgments}

We would like to thank the anonymous reviewers for their constructive comments. 
This work was supported by the SDAIA-KAUST Center of Excellence in Data Science and Artificial Intelligence (SDAIA-KAUST AI).
This publication is based upon work supported by the King Abdullah University of Science and Technology (KAUST) Office of Research Administration (ORA) under Award No FCC/1/1976-44-01, FCC/1/1976-45-01, URF/1/4663-01-01, RGC/3/4816-01-01, and BAS/1/1635-01-01.
This work was also supported by NSFC Grant No. 62122089 and CCF-Tencent Rhino-Bird Open Research Fund.

\bibliography{aaai23}

\end{document}